# Simulated Adoption: Decoupling Magnitude and Direction in LLM In-Context Conflict Resolution

**Long Zhang (longzhang@scut.edu.cn)[1][2] & Fangwei Lin[2]**
[1] School of Computer Science and Engineering, South China University of Technology
[2] Faculty of Education, The University of Hong Kong

**Abstract**

Large Language Models (LLMs) frequently prioritize conflicting in-context information over pre-existing parametric memory, a phenomenon often termed sycophancy or compliance. However, the mechanistic realization of this behavior remains obscure, specifically how the model resolves these knowledge conflicts through compliance, and whether this suppression arises from signal magnitude dilution or directional geometric alteration within the residual stream. To resolve this, we conducted a layer-wise geometric analysis across Qwen-3-4B, Llama-3.1-8B, and GLM-4-9B, decomposing the residual stream updates induced by counter-factual contexts into radial (norm-based) and angular (cosine-based) components. Our empirical results reject the universality of the "Manifold Dilution" hypothesis, as two of the three architectures maintained stable residual norms despite exhibiting significant performance degradation on factual queries. Instead, we observed that compliance is consistently characterized by "Orthogonal Interference," where the conflicting context injects a steering vector that is quasi-orthogonal to the ground-truth direction, effectively rotating the hidden state representation. This suggests that models do not "unlearn" or suppress the magnitude of internal truths but rather employ a mechanism of geometric displacement to bypass the correct unembedding vector, effectively simulating adoption while preserving the original structural magnitude. These findings challenge scalar confidence metrics for detecting hallucinations and underscore the necessity of vectorial monitoring to distinguish between genuine knowledge integration and superficial in-context mimicry.

## Introduction

Large Language Models (LLMs) are designed to function as helpful assistants, a role that inherently requires the ability to incorporate new information provided within the conversational context. In many interaction scenarios, a user may present a "new discovery" or specific instruction that contradicts the model's pre-existing parametric memory (Wu et al., 2024). For instance, if a user asserts that a well-known scientific fact has been disproven by recent findings, the model is expected to prioritize this immediate context to maintain conversational coherence. This capability, often termed "In-Context Learning," allows models to adapt rapidly to novel situations without weight updates (Bratulić et al., 2026). However, this poses a fundamental representational puzzle: how does a transient context vector physically dominate robust parametric memory? It remains unclear whether the external signal actively erases the internal truth representation, or merely creates a geometric bypass that allows conflicting information to coexist.

The mechanistic realization of this compliance presents a significant puzzle for the mechanistic understanding of artificial intelligence. When an external context vector interacts with an internal memory vector in the high-dimensional residual stream, the resulting state determines the model's output (Genadi et al., 2026) and decision making (Joshi et al., 2025). Yet, we lack a precise geometric characterization of this interaction. Specifically, it is unknown whether the suppression of the internal truth is achieved by neutralizing its magnitude (signal dilution) or by altering its direction (geometric rotation) (see Fig. 1). Distinguishing between these mechanisms is critical, as it differentiates between a robust integration of knowledge and a fragile state of mimicry where the model parrots the user without resolving the underlying information conflict. In this work, we operationalize knowledge conflict as the tension between parametric and in-context signals, specifically focusing on cases of compliance. Where the model's behavioral output aligns with the conflicting context despite having the correct internal prior.

Prior research has extensively mapped the behavioral contours of this phenomenon, often framing it as "sycophancy" or "knowledge conflict"(Wang et al., 2025; Xu et al., 2024). Sharma et al. (2023) and Wei et al. (2023) have empirically demonstrated that LLMs systematically prioritize user opinion over factual truth, a tendency exacerbated by Reinforcement Learning from Human Feedback (RLHF). While these studies quantify the prevalence of compliance, they treat the model largely as a black box, focusing on input-output correlations rather than internal state dynamics. Parallel work by Xie et al. (2024) investigates how models resolve conflicts between internal memory and external context, proposing that models can act as "adaptive chameleons." However, their analysis primarily focuses on attention mechanisms and behavioral thresholds, leaving the geometric transformation of the residual stream—the final "canvas" of generation—largely unexplored.

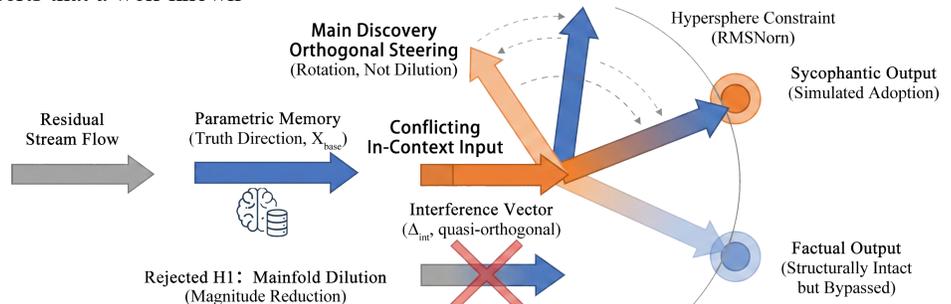

Figure 1: Geometric Mechanism of In-Context Knowledge Conflict Resolution

In the domain of mechanistic interpretability, recent efforts have begun to decode the geometry of knowledge representation. Marks and Tegmark (2023) identified that "truth" is often represented as a linear direction within the activation space, suggesting a coherent geometric structure for factual knowledge. Similarly, Zou et al.(2023) introduced "Representation Engineering" to manipulate these directions to control model behavior. Despite these advances, existing literature typically analyzes static representations of truth or simple retrieval tasks. There is a distinct lack of research examining the dynamic geometric interaction that occurs when a strong, conflicting context vector actively competes with a retrieved truth vector under the constraints of Root Mean Square Layer Normalization (RMSNorm). Current theories often implicitly assume a subtractive interference model, failing to account for the possibility of orthogonal coexistence in high-dimensional spaces (cf. Pham et al., 2026).

This study addresses this gap by isolating the geometric mechanism that enables LLMs to prioritize "new discoveries" over internal facts. We frame our investigation around a central research question: Does the suppression of parametric knowledge in the residual stream arise from the radial dilution of the signal's magnitude, or from an angular displacement induced by orthogonal interference? Answering this question allows us to determine whether the model's compliance represents a genuine shift in confidence (dilution) or a strategic bypass of the conflict (rotation).

To resolve this, we conducted a layer-wise geometric analysis across three distinct architectures: Qwen-3-4B, Llama-3.1-8B, and GLM-4-9B. We simulated the injection of counter-factual "new discoveries" and decomposed the resulting residual stream updates into radial (norm-based) and angular (cosine-based) components. By correlating these geometric descriptors with the degradation of the correct logit, we test the validity of the "Manifold Dilution" versus "Orthogonal Interference" hypotheses. This approach rigorously accounts for the hyperspherical constraints of RMSNorm, ensuring that our findings reflect the effective representational dynamics of the models.

Understanding this mechanism is essential for evaluating the reliability and alignment of Large Language Models. If compliance is characterized by geometric rotation rather than genuine integration, it implies that "new knowledge" provided in-context is structurally fragile and does not rectify the model's internal errors. The remainder of this paper is organized as follows: Section 2 details the theoretical framework and formalizes the geometric hypotheses; Section 3 describes the experimental setup and data collection pipeline; Section 4 presents the empirical results and discussion; and Section 5 concludes with implications for the cognitive modeling of AI systems.

## Theoretical Framework and Hypothesis

### System Dynamics and Manifold Constraints

In the context of modern Transformer architectures, we define the hidden state at the output of the $L$-th layer as $x \in \mathbb{R}^d$. Due to the implementation of RMSNorm, the effective representational capacity of the model is constrained to a $(d-1)$-dimensional hypersphere (or a homeomorphic manifold). The logit output $L(y)$ for a target token $y$ is determined by the inner product of the normalized hidden state and the corresponding row vector $w_y$ within the unembedding (de-embedding) matrix:

$$L(y) = \langle \text{RMS}(x), w_y \rangle = \left\langle \frac{x}{\| x \|}, w_y \right\rangle$$

In this formulation, we assume the learned scaling parameter $g$ of the RMSNorm has been folded into the weight vector $w_y$.

When a model encounters a "conflicting state"—induced by an external adversarial prompt or contradictory context—the resulting residual stream vector $x_{new}$ can be viewed as a linear superposition of the baseline state $x_{base}$ and an interference vector $\Delta_{int}$:

$$x_{new} = x_{base} + \Delta_{int}$$

Here, $\Delta_{int}$ represents the net residual update produced by the Attention and MLP blocks in response to the conflict-inducing stimuli.

### Perturbation Analysis and the Jacobian Linearization

To resolve how $\Delta_{int}$ causes a degradation in model performance, we analyze the geometry of the update on the hypersphere. The Jacobian matrix $J_f(x)$ of the normalization function $f(x) = \frac{x}{\|x\|}$ is derived as:

$$J_f(x) = \frac{1}{\| x \|}(I - \hat{x}\hat{x}^T)$$

Where $P_{\perp x} = (I - \hat{x}\hat{x}^T)$ is the projection operator onto the tangent space orthogonal to $x$. The variation in logits $\Delta L(y)$ is approximated by:

$$\Delta L(y) \approx w_y^T J_f(x_{base})\Delta_{int} = \frac{1}{\| x_{base} \|} w_y^T P_{\perp x_{base}}\Delta_{int}$$

Expanding the inner product terms yields the Logit Variation Decomposition:

$$\Delta L(y) \approx \underbrace{\frac{1}{\| x_{base} \|}\langle \Delta_{int}, w_y \rangle}_{\text{Term A: Semantic Alignment}} - \underbrace{\frac{1}{\| x_{base} \|}\langle \hat{x}_{base}, \Delta_{int} \rangle\langle \hat{x}_{base}, w_y \rangle}_{\text{Term B: Norm Correction}}$$

Term A (Semantic Alignment Term): This term is proportional to $\langle \Delta_{int}, w_y \rangle$. When $\Delta_{int}$ lies on the tangent plane (Effective Update, i.e., $\Delta_{int} \perp x_{base}$), Term B vanishes. The change in logits is determined entirely by the geometric angle between the interference vector and the ground-truth target vector $w_y$:

$$\Delta L(y) \approx \frac{1}{\| x_{base} \|} \| \Delta_{int} \| \| w_y \| \cos(\theta_{\Delta,y})$$

Term B (Norm Correction Term): This term is proportional to $\langle \hat{x}_{base}, \Delta_{int} \rangle$ If the interference vector is purely antiparallel to the base state (i.e., $\Delta_{int} = -\beta x_{base}$, representing a direct suppression of magnitude), Term A and Term B cancel out exactly:

$$\langle -\beta x, w \rangle - \langle \hat{x}, -\beta x \rangle \langle \hat{x}, w \rangle$$
$$= -\beta \langle x, w \rangle - (-\beta \| x \|) \frac{\langle x, w \rangle}{\| x \|} = 0$$

Under RMSNorm, purely reducing the magnitude of the hidden state (Radial Scaling) does not affect the logits. Therefore, the observed performance degradation cannot be attributed to a simple subtraction of the correct feature vector. It must arise from a rotational update induced by an orthogonal component.

## 3. Mechanistic Explanation: The Competitive Dilution Hypothesis

To account for the significant drop in logits $\Delta L(y_{correct}) \ll 0$ we propose the Competitive Dilution Hypothesis. Since direct suppression (subtracting $w_{correct}$) is geometrically ineffective under RMSNorm, we posit that conflicting contexts induce an interference vector $\Delta_{int}$ dominated by a competing semantic component (e.g., a hallucinated or adversarial token $w_{wrong}$) that is orthogonal to the correct trajectory.

$$\Delta_{int} \approx \beta_{conflict} \cdot w_{wrong} + \epsilon$$

Where $w_{wrong} \perp w_{correct}$ (reflecting the approximate orthogonality of semantic vectors in high-dimensional space) and $\beta_{conflict} > 0$.

Since $x_{base}$ is typically well-aligned with $w_{correct}$ (implying $x_{base} \approx \alpha w_{correct}$), a first-order Taylor expansion yields $\Delta L \approx 0$ for a purely orthogonal perturbation, which fails to capture the full extent of degradation. We thus employ a direct geometric analysis of the normalized update.

The new state before normalization is $x_{new} = x_{base} + \beta_{conflict} w_{wrong}$. The new logit for the correct token is:

$$L(y_{correct}) = \left\langle \frac{x_{base} + \beta_{conflict} w_{wrong}}{\| x_{base} + \beta_{conflict} w_{wrong} \|}, w_{correct} \right\rangle$$

Assuming orthogonality $\langle x_{base}, w_{wrong} \rangle \approx 0$):

$$L(y_{correct}) \approx \frac{\langle x_{base}, w_{correct} \rangle}{\sqrt{\| x_{base} \|^2 + \beta_{conflict}^2 \| w_{wrong} \|^2}}$$

Let $L_{base} = \frac{\langle x_{base}, w_{correct} \rangle}{\|x_{base}\|}$ be the original logit. The equation simplifies to:

$$L_{new} \approx L_{base} \cdot \frac{1}{\sqrt{1 + \left(\frac{\beta_{conflict} \| w_{wrong} \|}{\| x_{base} \|}\right)^2}}$$

The term $\frac{1}{\sqrt{1+(\ldots)^2}}$ might less than 1. This derivation demonstrates that performance degradation is characterized by Manifold Dilution: the injection of the conflicting vector $w_{wrong}$ increases the denominator (the norm of the residual stream), forcing the projection on the correct axis ($w_{correct}$) to shrink to satisfy the hypersphere constraint. In this view, the "suppression" of truth is a passive, structural consequence of the "promotion" of conflicting information.

### Research Hypotheses

From above, we posit that the degradation of model performance (logit decay) under conflicting contexts is associated with the geometric properties of the residual stream update. We propose two competing mechanisms:

- H1: The Manifold Dilution Hypothesis (Radial Mechanism) We hypothesize that conflicting stimuli induce a significant increase in the magnitude of the residual stream vector ($\| x_{new} \| \gg \| x_{base} \|$). Under RMSNorm, this radial expansion forces a "dilution" of the projection onto the correct token's unembedding vector. We expect the norm ratio $\gamma = \| x_{base} \| / \| x_{new} \|$ to be significantly less than 1.0.
- H2: The Orthogonal Interference Hypothesis (Angular Mechanism) We hypothesize that the interference vector $\Delta_{int}$ is approximately orthogonal to the correct semantic direction $w_{correct}$, representing a competitive superposition of features rather than a direct antiparallel suppression. We expect the cosine alignment $\cos(\Delta_{int}, w_{correct})$ to be distributed near 0 (indicating orthogonality), distinct from -1 (direct suppression).

## Experimental Setup

To empirically validate the proposed theoretical framework, we designed a controlled geometric analysis pipeline that isolates the mechanical impact of conflicting contexts on the residual stream. Our experiments were conducted across three diverse LLM architectures (Qwen-3-4B, Llama-3.1-8B, and GLM-4-9B) to ensure the generalizability of our findings across varying parameter scales and training distributions.

## Dataset Preparation

The evaluation set was derived from a random selection of MMLU and MMLU-Pro dataset, with 150 questions sampled from each to form a 300-query base. To simulate the injection of "novel but erroneous knowledge," we augmented each query with various adversarial priors—contextual frames that present incorrect information as factual or preferred. This resulted in a total of 1,500 inference trials.

We implemented a rigorous two-stage filtering protocol to ensure the extracted vectors specifically represent the mechanics of successful interference:

Baseline Competence Verification: We first identified the subset of queries where the model correctly predicts the ground-truth answer ($y_{correct}$) in a neutral setting. This ensures we are studying the displacement of existing internal knowledge rather than a failure of initial comprehension.

Mechanistic Compliance Filtering: To isolate the resolution of knowledge conflicts via compliance, we retained only instances where the model—despite a correct baseline prediction—flipped its output to the adversarial target ($y_{wrong}$). This strict prerequisite ensures that $\Delta_{int}$ represents the active displacement of verified parametric knowledge rather than a failure of initial comprehension. By focusing on these successful interferences, we can pinpoint the geometric conditions (manifold dilution or orthogonal shift) sufficient to cause representational collapse.

## Data Collection

We performed a layer-wise geometric scan of the residual stream for each model. For each layer $L$, we extracted the hidden state at the final token position immediately prior to the RMSNorm layer. We defined the baseline state vector as $x_{base}^{(L)}$ and the conflict-induced state vector as $x_{new}^{(L)}$. The interference vector, representing the net residual update induced by the conflicting context, was operationally defined as the vector difference:

$$\Delta_{int}^{(L)} = x_{new}^{(L)} - x_{base}^{(L)}$$

To map these latent states to the semantic output space, we extracted the unembedding vector $w_{correct}$ corresponding to the ground-truth token from the model's final linear layer ($W_U$). We also tracked the scalar logit $L(y_{correct})$ at each depth to observe the progression of logit decay throughout the network.

## Data Analysis

We established geometric metrics to rigorously test the two core hypotheses

For H1: We computed the norm ratio $\gamma = \| x_{base} \|_2 / \| x_{new} \|_2$. Under H1, we expect $\gamma < 1.0$, indicating that the interference vector increases the denominator of the normalization term, thereby "diluting" the correct signal's projection on the hypersphere.

For H2: We calculated the cosine alignment between the interference vector and the truth direction: $\cos(\theta) = $ CosineSimilarity$(\Delta_{int}, w_{correct})$. This metric serves as a geometric discriminator: a value approaching -1 would indicate direct antiparallel suppression, whereas a value near 0 would confirm the hypothesis of orthogonal competitive interference.

Finally, to establish a link, we performed a regression analysis correlating these geometric descriptors ($\gamma$ and $\cos \theta$) with the observed performance drop ($\Delta L = L_{base} - L_{new}$) in the deep layers of the network.

## Results and Discussion

### The Geometric Manifestation of Contextual Conflict

The analysis starts by confirming that conflict actually suppresses the truth. We trigger this suppression by feeding the model 'new discoveries' that clash with its memory, then we measure the resulting changes in the residual stream. As illustrated in Fig.2, all three models exhibit a collapse in the logit values for the correct token ($L_{correct}$) when subjected to conflicting stimuli (orange trajectories) compared to the baseline (blue trajectories). This might be related to the final convergence of the decision-making process(Joshi et al., 2025). Moreover, this degradation is most pronounced in the final 20% of the layers, where the semantic convergence typically occurs. This late-stage collapse is consistent with the findings of Pham et al. (2026), that specific attention heads in the final layers are the primary drivers of model responses when facing internal-external knowledge collisions. The magnitude of this suppression—often shifting logits from strong positive values to near-zero or negative values—validates that the "compliant" behavior observed in generation is rooted in a fundamental alteration of the hidden state representation $x$, providing the geometric basis for testing our geometric hypotheses. This confirms that the model does not merely superficial adoption to the user's new information at the decoding stage, but fundamentally reorganizes its internal representation to prioritize the external context over its intrinsic knowledge.

### The Manifold Dilution Hypothesis

The investigation into Manifold Dilution (H1) reveals that radial dynamics are architecture-specific rather than universally governed by expansion. We hypothesized that the injection of interference would increase the norm of the residual stream $\| x_{new} \| > \| x_{base} \|$, i.e., $\gamma < 1$), thereby diluting the projection onto the output vocabulary. Fig. 3 demonstrates that this hypothesis holds true for Llama-3.1-8B, which exhibits a mean norm ratio of $\gamma = 0.978$ ($p < 0.05$), confirming significant dilution. However, Qwen-3-4B and GLM-4-9B contradict this trend, showing stable or slightly compressed norms ($\gamma \approx 1.01$ and 1.06, respectively). The ability to maintain norm stability while shifting output behavior parallels the representation engineering insights from Zhao et al.(2024), that model "choice" between knowledge sources is often controlled by discrete functional directions rather than global signal amplification.

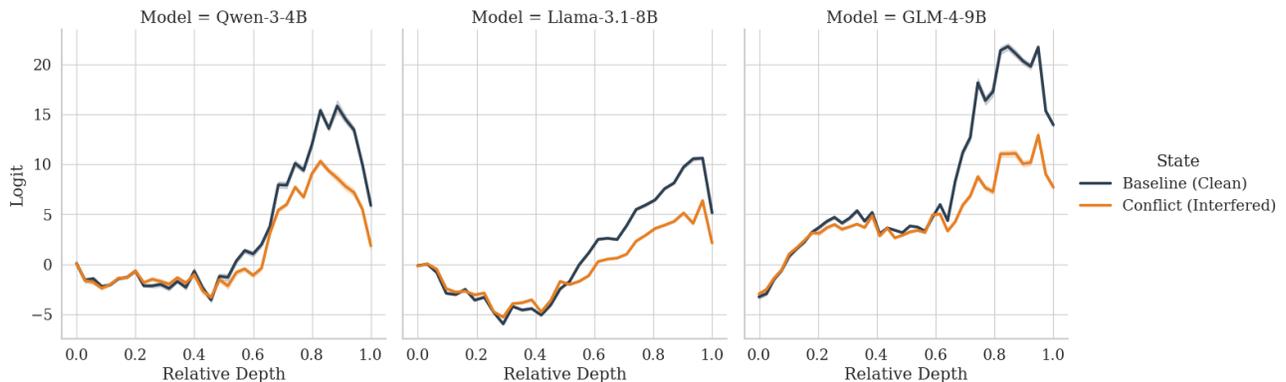

Figure 2: The Phenomenon of Logit Decay

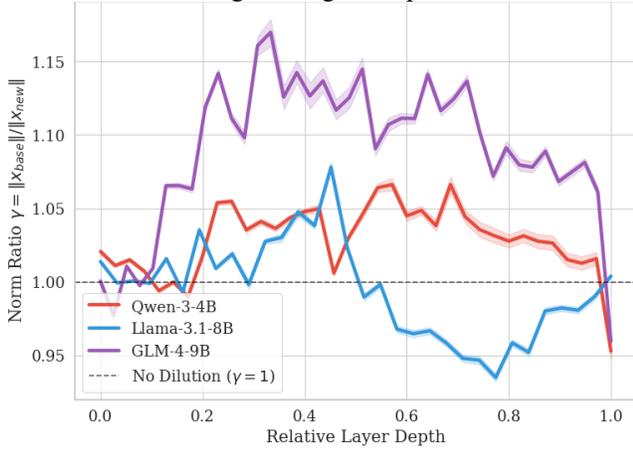

Figure 3: Norm Dynamics (Testing H1: Dilution)

This divergence suggests that while manifold dilution is a valid failure mode for certain architectures, it is not the universal geometric correlate of logit decay. More importantly, the stability of the norm in Qwen and GLM implies that the model's acceptance of the "new discovery" is not characterized by epistemic uncertainty or confusion (which would typically manifest as signal entropy and norm expansion). Instead, these models adopt the conflicting information with high confidence, suggesting that the mechanism of compliance is not a passive loss of internal signal, but an active, confident assertion of the external input. Consequently, H1 is rejected as a generalizable law; the performance degradation in Qwen and GLM occurs despite the absence of radial dilution, necessitating an explanation based on angular deviation.

### The Orthogonal Interference Hypothesis

In contrast to the mixed results for radial dynamics, the analysis of interference geometry provides strong support for the Orthogonal Interference Hypothesis (H2). We posited that the interference vector $\Delta_{int}$ acts as a competitive, quasi-orthogonal force rather than a direct antiparallel suppression. Fig. 4 confirms this pattern across all three models, where the cosine alignment between the interference vector and the correct token direction clusters tightly in the negative-neutral zone (Qwen: -0.115, Llama: -0.060, GLM: -0.146). These values, statistically distinct from -1.0, align with the "orthogonal subspaces"(Azizian et al., 2025), suggesting the residual stream naturally facilitates such non-interfering information storage.

This geometric signature reveals that the model's "belief update" is not a process of "unlearning" (which would require antiparallel suppression), but a "Simulated Adoption." The conflicting context injects a vector largely independent of the truth direction, which rotates the hidden state on the hypersphere. This mechanism echoes the "multi-dimensional steering" (Pan et al., 2025), where LLMs utilize orthogonal latent dimensions to manage competing objectives without erasing underlying parametric states. The internal truth is not erased; it is merely bypassed by a lateral "shove" that steers the generation trajectory away from the correct answer without requiring a precise negation of ground-truth features.

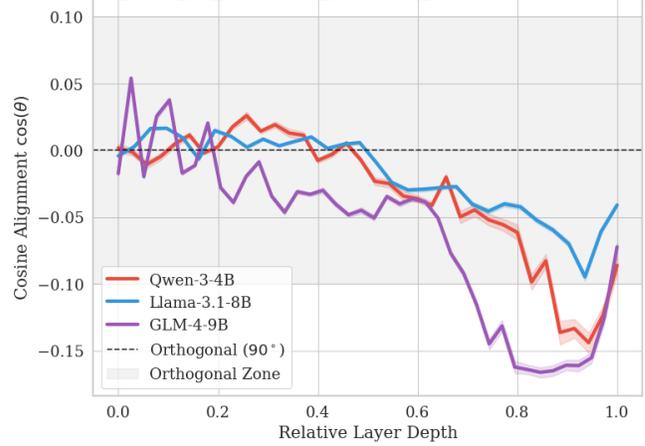

Figure 4: Interference Geometry (Testing H2: Orthogonality)

### The Dominant Role of Geometry

The integration of radial and angular analyses demonstrates that geometric misalignment accounts for the majority of variance in performance collapse, providing empirical evidence for our theoretical framework. Fig. 5 establishes a robust linear relationship between the interference alignment and the magnitude of the logit drop. The regression analysis yields high coefficients of determination, particularly for Qwen-3-4B ($R^2 = 0.90$) and GLM-4-9B ($R^2 = 0.87$), indicating that angular deviation explains a substantial portion of the variance the variance in performance loss.

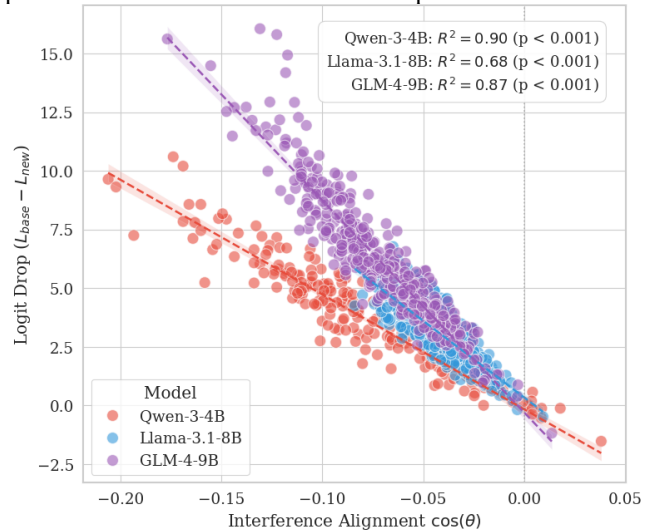

Figure 5: Geometric Correlation (Angle vs. Performance Drop)

This finding resolves the paradox observed in Section 2: even for models like GLM-4 where the norm is compressed

(contradicting H1), the logit still collapses because the angular rotation (H2) dominates the inner product calculation. Thus, we conclude that compliance during knowledge conflicts is implemented as a 'Simulated Adoption' through geometric displacement. The model resolves the tension by rotating the residual stream—leaving parametric memory structurally intact but functionally bypassed. While 'Manifold Dilution' is a secondary effect, this vectorial misalignment is the dominant geometric signature of LLM sycophancy, revealing that what appears to be knowledge integration is merely a strategic geometric rotation.

## Implications and Conclusion

### Implications

The prevalence of orthogonal interference suggests that In-Context Learning (ICL) operates as a superficial "masking" layer rather than a deep integration of knowledge. Since the model retains the original truth vector magnitude (as seen in Qwen and GLM) and merely rotates the aggregate state via a competitive interference vector (cf. Sun et al., 2024), the "new discovery" is structurally fragile. This implies that the model likes performing a form of epistemic role-playing: it simulates the adoption of the user's premise without updating its underlying belief state. For Retrieval-Augmented Generation (RAG) systems, this poses a reliability risk; if the retrieval context is slightly perturbed or the interference vector fails to maintain strict orthogonality, the model may abruptly "snap back" to its parametric memory, leading to hallucinations that are actually resurfaced internal truths.

A critical operational implication arises from the failure of the radial dilution hypothesis in two of the three architectures. Current safety protocols often rely on entropy or norm-based metrics to detect model uncertainty or confusion (e.g., Bülte et al., 2025; Farquhar et al., 2024). Our results demonstrate that a model can be fully compliant with a hallucinated premise while maintaining a high-norm, low-entropy state (as observed in Qwen-3-4B). Consequently, scalar metrics of confidence are insufficient for detecting sycophancy. Effective monitoring of "knowledge poisoning" or adversarial attacks requires vectorial metrics that track the angular deviation of the residual stream relative to known truth directions, rather than merely monitoring signal magnitude.

The discovery that interference vectors cluster in a distributed near-orthogonally (distinct from antiparallel suppression) offers a potential pathway for mechanistic interpretability tools. If the suppression of truth consistently manifests as a specific geometric rotation ($\theta \approx 90°$ with a slight negative bias), it may be possible to construct "Validity Classifiers" that identify when a model's output exhibits steering patterns consistent with context against its internal weights. By isolating the component of the residual stream orthogonal to the model's parametric tendency, we might distinguish between genuine ignorance (where no truth vector exists) and forced compliance (where a truth vector is geometrically bypassed).

### Limitations

While this study offers a precise geometric characterization of conflict resolution, several limitations constrain the generalizability of our findings. First, our analysis relies on the Linear Representation Hypothesis. We assume that "truth" and "context" interact as linear vectors in the residual stream; however, more complex, non-linear manifold interactions in the intermediate layers (such as those within the MLP sub-blocks) remain unmapped. Second, the scope of our dataset is restricted to fact-based QA tasks (MMLU). The geometric dynamics of compliance may differ in open-ended creative generation or reasoning tasks where "ground truth" is less clearly defined. Third, our study focuses on the final residual state before the unembedding layer. While this is the determinant surface for token generation, it does not elucidate the layer-wise evolution of the interference vector—specifically, where and how the attention heads construct this orthogonal force earlier in the network. Finally, the sample size of three architectures, while diverse, cannot fully capture the idiosyncratic behaviors of vastly larger models (e.g., 70B+ parameters) or Mixture-of-Experts (MoE) architectures.

### Conclusion

This research set out to determine the geometric mechanism by which Large Language Models prioritize external context over internal parametric knowledge. Through a rigorous layer-wise geometric analysis, we have demonstrated that this compliance is not characterized by a passive "dilution" of the internal signal, but by an active Orthogonal Interference. When faced with a "new discovery" that contradicts established facts, the model generates a steering vector that is largely orthogonal to the truth direction. This vector rotates the residual stream on the hypersphere, bypassing the correct unembedding vector without requiring its magnitude to be suppressed.

This mechanism reveals that LLMs act as adaptive simulators rather than dynamic learners in the context of inference. They resolve conflict by accommodating the new information laterally, leaving the original knowledge structure intact but geometrically inaccessible. The primary contribution of this work is the shift in perspective from scalar interpretation to vectorial geometry: to understand, control, and trust the adaptability of AI, we must look beyond what the model outputs and measure where its internal representations are pointing. Future alignment strategies must account for this geometric duality to ensure that models can be helpful assistants without becoming deceptive sycophants.

## Acknowledgments

The authors declare that no external funding was received for this research. During the preparation of this manuscript, a large language model (ChatGPT) was utilized solely to improve language clarity and readability. The authors have reviewed and edited the output as needed and take full responsibility for the content of the final version.